# The Role of Pragmatics in Legal Norm Representation

Shashishekar Ramakrishna[1], Łukasz Górski[2] and Adrian Paschke[1]

[1] Freie Universität Berlin, Berlin, Germany.
{shashi792@gmail.com; paschke@inf.fu-berlin.de}
[2] Nicolaus Copernicus University, Toruń Poland.
{lukasz.gorski@gmail.com}

**Abstract.** Despite the *'apparent clarity'* of a given legal provision, its application may result in an outcome that does not exactly conform to the semantic level of a statute. The vagueness within a legal text is induced intentionally to accommodate all possible scenarios under which such norms should be applied, thus making the role of pragmatics an important aspect also in the representation of a legal norm and reasoning on top of it. The notion of pragmatics considered in this paper does not focus on the aspects associated with judicial decision making. The paper aims to shed light on the aspects of pragmatics in legal linguistics, mainly focusing on the domain of patent law, only from a knowledge representation perspective. The philosophical discussions presented in this paper are grounded based on the legal theories from Grice and Marmor.

## 1 Introduction

The three major building block of a language are its *Syntax, Semantics* and *Pragmatics*. While, the *syntax* defines the grammar and punctuation of the language, *Semantics* provides the meaning of constructs in a language and *Pragmatics* has to do with how well the language connects to the 'real world'.

The role of *Syntax, Semantics* and *Pragmatics* in a language can be easily understood with an example. Consider a "Traffic-Light Situation Awareness" scenario, where the lights displayed are of standard colors (red, yellow, and green) following a universal color code. Wherein the:

- **Syntax:** Bottom light - Green, Middle light - Yellow and Top Light - Red
- **Semantics:** Green light allows traffic to proceed in the direction denoted, Yellow light provides warning that the signal will be changing from green to red and red light prohibits any traffic from proceeding.
- **Pragmatics:** You are obliged to stop at a green light and make way for an approaching emergency service vehicle.

This paper contributes in discussing the notion of pragmatics and how it enriches the law, as the importance of adherence to syntax and semantics in the legal context. The notion of *'pragmatics'* considered in this paper does not focus on the aspects associated with judicial decision making but only those needed for legal norm representation. In particular, the aspects from the patent law which are expressed through its manuals (e.g. MPEP or MPPP) describing the procedures used for applying the respective substantive laws (e.g. § 112 of US patent law or § 2(1)(ja) of Indian patent law). However, its premises can be applied *mutatis mutandis* to other kinds of legal discourse. The philosophical discussions presented in this paper are grounded based on the legal theories from Grice [1], Habermas [2] and Marmor [3]. The paper addresses these issues from a knowledge modeling perspective.

The paper is structured as follows: Section 2 discusses the new notion *active web* by introducing the concept of Pragmatic Web 4.0. Section 3 addresses the interconnection between legal linguistic theory and speech-act theory and the aspect of pragmatics in statutory laws. Section 5 with the help of maxims, provides a solution for interpretation of legal norms with their intended pragmatics and how such interpretation with associated pragmatics are/can be captured for knowledge modeling and reasoning on top of such modeled legal norms for its further use such in; decision systems, court-filings, in-court argumentation etc.. Section 6 provides conclusion and future works.

## 2 The Pragmatic Web

Weigand et.al [4] defines The Pragmatic Web [5] as a particular view on the internet as a platform of communication and coordination [6]. Pragmatic Web consists of the tools, practices and theories describing *Why* and *How* people use information. Compared to the *Syntactic Web* (e.g. Webpages linked by HTML references) which is about form or syntactic information resources and the *Semantic Web* (e.g. Ontologies) which is about the semantic resources (meaning) concerning the Syntactic Web, the *Pragmatic Web* is about the interaction which brings about e.g. understanding or commitments [7]. The *Pragmatic Web* enriches the *Semantic Web* by considering the necessary context information and lets this context information to evolve.

Norms are represented and reasoned using rule-based systems. Such systems encompass both rule representation languages such as RuleML [8], CARIN [9], languages specifically designed to handle legal rules like LegalRuleML [10], KRIP [11], KR4IPLaw [12] etc.. and Rule Reasoners such as Drools, Prova [13] etc..

Rules as such provide a mechanism to control and reuse the manifold semantically linked meaning representations published on the semantic web. The inclusion of the underlying pragmatics (e.g. in the form of micro-ontologies) allows for automated intelligence, which goes beyond simple deductive reasoning. According to Weigand and Paschke [7], an envisioned ubiquitous Pragmatic Web 4.0 - an extension of current Semantic Web 3.0 - is as depicted in Fig 1

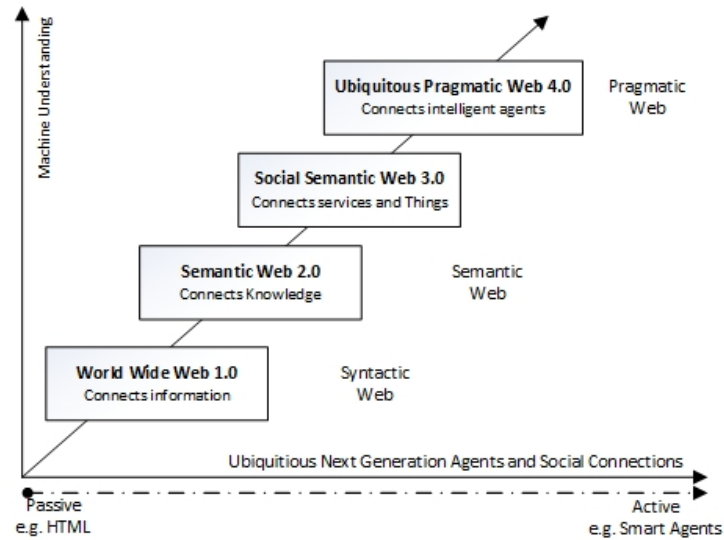

Fig. 1: Envisioned Pragmatic Web 4.0 (adapted from [7])

## 3 Pragmatics in Law

The distinction of syntax, semantics and pragmatics applies to law as well. Though the linguistic analysis of the legal language has not been the subject of very extensive studies, a number of researchers have already suggested that law could be viewed using the conceptual framework offered, *inter alia* by the speech act theory. In this context, Habermas' theory of communicative action is a shining example [2][14].

The basic elements of linguistic conceptual framework of syntax, semantics and pragmatics applies to law as well as any other act of communication [15]. All three facets of the language gain an importance especially in the context of legal argumentation [16].

Despite the apparent clarity of a given legal provision, its application to a particular case may result in an outcome that does not exactly conform to the semantic level of a statute. There may be other conflicting laws in place, and it is a matter of interpretation to choose the prevailing one. Additionally, application of a particular provision may result in an outcome that is absurd or unjust, so the provision has to be interpreted in a way that yields more acceptable results [17]. One should also account for the fact that legislature usually passes a law with a purpose, and that it may be taken into account when interpreting statutes. Holmes also pointed out the importance of the context: "If Congress has been accustomed to use a certain phrase with a more limited meaning that might be attributed to it by common practice, it would be arbitrary to refuse to consider that fact when we come to interpret the statute" [15]. Similar assumption was put forth by Marmor: "[T]he law typically speaks to a legal community, not to lay people. Legislatures in a modern society mostly address their laws to legal experts, such as judges, lawyers, administrative agencies, etc. There is typically a legal culture that would share a great deal of information and contextual knowledge that is much greater than the relative contextual knowledge of the population at large." [17]

### 3.1 Speech Theory and Patent Law

Speech-act theory is based on conversations to understand the content and the context behind a utterance. Statutes are a way of passing information (e.g. requirements or constraints) between a person/body responsible for creating such statutes and its intended subjects (generally bound by a jurisdiction). This analogy provides us with a conceptual framework for applying the concepts of speech-act theory to legal linguistics theory dealing with the interpretations of these statutes.

This relation between contextual information and language has already been carefully studied on the field of linguistics, most famously by Grice [3][15][18]. He viewed the conversation as the cooperative endeavor, where not only what has been explicitly said, but also what has been tacitly assumed and is known to both conversation parties matter. Such conversation is guided by the following conversational maxims, which - while obeyed or deliberately flouted - allow to decode what the utterance means on pragmatic level:

- **Maxim of Relevance** - the conversational utterance has to make a contribution to the conversation somehow; i.e. when a conversation party answers that "weather is bad" to a suggestion for going outside, the relevancy maxim says that most probably he meant to decline the proposal and not only comment on a weather.
- **Maxim of Quantity** - the conversational contribution has to be precisely as informative as required; it should not state more nor less than the speaker intends to; such maxim would be violated in the following sample exchange: A: Can you tell me the time? B: Yes.
- **Maxim of Quality** - don't say something you don't believe to be true; sample exchange that violates this maxim: A: What are major building blocks of a language? B: Only semantics and syntactics.
- **Maxim of Manner** - be brief and orderly; any overly long and complicated utterance violates this maxim.

In the domain of intellectual property laws (including patent laws), as described by Osenga [18], the process of applying the substantive laws on inventions/patent applications can be viewed as a non-verbal conversation between the legislature and patent officer/examiner, and further- between the inventor and the patent office/examiner. Such conversation can be seen in similar lines to utterances as defined by the speech-act theory. According to Marmor, the norms of the statutory interpretation that are adopted by the courts over time can be understood as a sort of Gricean maxims of conversation in legal context. However, it is more problematic to define their scope and exact content [17]. The following examples show how the pragmatics enriches the semantics of legal provision and how the courts and patent office seek to look for a context that allows to understand the legal provisions regarding the patent law.

Such enrichment can be seen, for example, in the court-mandated doctrine of equivalents (see, for example, *Graver Tank and Manufacturing Co. v. Linde Air Products Co.* [19] or *Winans v. Denmead* [20]). In those cases the courts have found that "In determining whether an accused device or composition infringes a valid patent, resort must be had in the first instance to the words of the claim. If accused matter falls clearly within the claim, infringement is made out, and that is the end of it" [19]. However, taking into account only the literal sense of the patent would be void as it would allow "unscrupulous copyist to make unimportant and insubstantial changes and substitutions in the patent which, though adding nothing, would be enough to take the copied matter outside the claim, and hence outside the reach of law" [19]. Hence, such interpretation would allow to upkeep the patent law's main purpose - protection of inventor and his monopoly over the invention. Doctrine of equivalents is not codified in statutory law, therefore can not be constructed on the basis of the semantics of legal provisions related to patents - it constitutes the pragmatic aspect of patent law.

In its decision in *Ex parte Lundgren*, the USPTO implicitly recalled the maxim of quantity. In its interpretation of § 101 it stated that does not pertain only to "technological arts", as no such reservation was made therein. Obviously, such analysis was also supported by invoking the other aspects of pragmatics, as USPTO inspected a number of prior judgments pertaining to § 101 to assure that "technological art test" has not been invoked by the court before.

## 4 Formal Pragmatics

If $\mathcal{D}$ is a domain model of the language under discussion, $\alpha$ a legal text/paragraph from the domain $\mathcal{D}$, a set of questions on $\alpha$ can be written as:

$$q(\alpha)_{n,m} = \{[(p(\alpha)_1, wh_1)......(p(\alpha)_1, wh_m)].....[(p(\alpha)_n, wh_1)......(p(\alpha)_n, wh_m)]\}$$

Wherein, $p(\alpha)_i, 1 \leq i \leq n$ are the set of propositions on $\alpha$ and $wh_j, 1 \leq j \leq m$ are the set of *wh*-questions i.e. 'who', 'what' etc.... The validity of a *wh*-question depends on their acceptance by the interlocutors.

An interpretation of a legal text/paragraph $\alpha$, $\mathcal{I}(\alpha)$ may be defined as a set of $\mathcal{A}(\alpha, q)$, where $\mathcal{A}$ is an answer to a *wh*-question on the legal text/paragraph $\alpha$.

**Definition 1. Pragmatics,** $\mathcal{P}$ *associated with a legal text/norm, $\mathcal{P}(\alpha)$ may be defined as a set of interpretations or utterances (according to speech-act theory) about a norm.*

$$\mathcal{P}(\alpha) = \{\mathcal{I}_1(\alpha), \mathcal{I}_2(\alpha)...., \mathcal{I}_n(\alpha)\} \,|\mathcal{I}_i(\alpha) \in \mathcal{D} \text{ and } 1 \leq i \leq n.$$

Wherein each interpretation, $\mathcal{I}_i(\alpha)$ provides a part of the context information associated to the content i.e. legal norm.

**Definition 2. Elementary Pragmatics,** $\mathcal{EP}(\alpha)$ *is a part of the context information (either implicit, explicit, intrinsic or extrinsic) associated with a legal norm* $\alpha$. $\mathcal{EP}(\alpha)$ *may be defined as below:*

$$\mathcal{EP}(\alpha) = \mathcal{A}((p(\alpha)_i, wh_j))|\ 1 \leq i \leq n.\text{and}\ 1 \leq j \leq m.$$

A combination of all context information provide all the necessary pragmatics required for modeling a legal norm for its use in (semi-)automated reasoning.

## 5 Interpretation of Pragmatics

As discussed before, statues can be seen as an non-verbal textual unilateral communication between the legislative body responsible for creating such statutes and its intended subjects (generally bound by a jurisdiction). In the domain of patent laws (Title 35 of the United States Code), the legislative/congress frames the substantive patent laws for its use by the USPTO - inturn by an patent examiner- for examining a patent application for it satisfying the patent norms [3].

Due to the obvious differences between everyday's conversation and the "legislature speech", Grice's maxims do not always apply to legal provisions in their original form. Sinclair [15] extended these maxims to connect with well-established canons of statutory interpretation. We further adapt the maxims to the patent law domain, thereby, providing an qualitative approach for non-ambiguous interpretation of substantive patent norms, which further can be modeled for (semi-/)-automated reasoning.

- **Maxim of Relevance** - The scope of the statute defines the boundary of each provision and its interpretation; i.e. while US copyright law has provisions that pertain to the works made under the employment contract ("works made for hire"), the patent law does not and copyright law provisions shouldn't be extended to patents.
- **Maxim of Quantity** - A statutory provision does not apply to entities or behaviors which are not in its specific domain and does not place controls on entities or behaviors beyond those specified; the *Ex parte Lundgren* mentioned hereinbefore is of an example.
- **Maxim of Quality** - Every provision should be interpreted in the light of its objective; the already *Graver Tank and Manufacturing Co. v. Linde Air Products Co.* case can be invoked as an example in this respect.

---

[3] The other node of the patent examination communication (textual & argumentative) between the applicant and the USPTO, in the form of 'Office Action Response', is out of scope for this paper

– **Maxim of Manner** - The statute should be interpreted according to plain, ordinary meaning of its provisions (i.e. in line with the doctrine of plain ordinary meaning), unless explicitly stated otherwise; 35 USC 100 enumerates a number of definitions that alter the plain meaning of a number of terms - i.e. it states that patentee includes not only the patentee to whom the patent was issued but also the successors in title to the patentee.

Capturing the pragmatics associated to a norm in-line to the maxims is an important step in the direction of modeling these norms.

To illustrate the identification, interpretation and representation of a legal norm with its associated pragmatics, consider Paragraph 1 of Section 112, of the United States patent law, dealing with the patent enablement.

*35 U.S. Code § 112 - Specification*
*(a) In General –The specification shall contain a written description of the invention, and of the manner and process of making and using it, in such full, clear, concise, and exact terms as to enable any person skilled in the art to which it pertains, or with which it is most nearly connected, to make and use the same, and shall set forth the best mode contemplated by the inventor or joint inventor of carrying out the invention.*

The legal section is disaggregated from its vague substantive law semantics into a concrete procedural (norm) semantics, to extract the elementary concerns from the compound concerns of the statutory law. In the US patent law, the United States Patent and Trademark Office (USPTO), through its Manual for Patent Examination Procedure (MPEP), provides to its examiners, such elementary concerns in the form of procedural laws. In-line to the maxims discussed earlier, we transform such procedures into legal decision models, wherein, each decision point is a single procedure or a set of procedures to be carried out. The landmark decisions (*In re Ruschig Fed Cir* and *Pfizer Inc. v. Teva Pharmaceuticals Inc*) are also represented as procedural decision models and are integrated into the decision model generated for the specific legal section. Thus forming a generic decision model for a specific legal section as shown in Fig 2 [4]. This allows capturing the different interpretations of the same section in different case laws. A more in-depth discussion pertaining to the legal decision model has been presented by us in [21]

*Procedural Norm (decision point 'D') i.e. MPEP ¶ 7.33.01*
*Claim [1] rejected under 35 U.S.C. 112, first paragraph, as based on a disclosure which is not enabling. [2] critical or essential to the practice of the invention, but not included in the claim(s) is not enabled by the disclosure*
  *1. This rejection must be preceded by form paragraph 7.30.01 or 7.103.*
  *2. In bracket 2, recite the subject matter omitted from the claims.*

---
[4] Note: The textual content inside the decision model is left out on purpose to handle the space restrictions

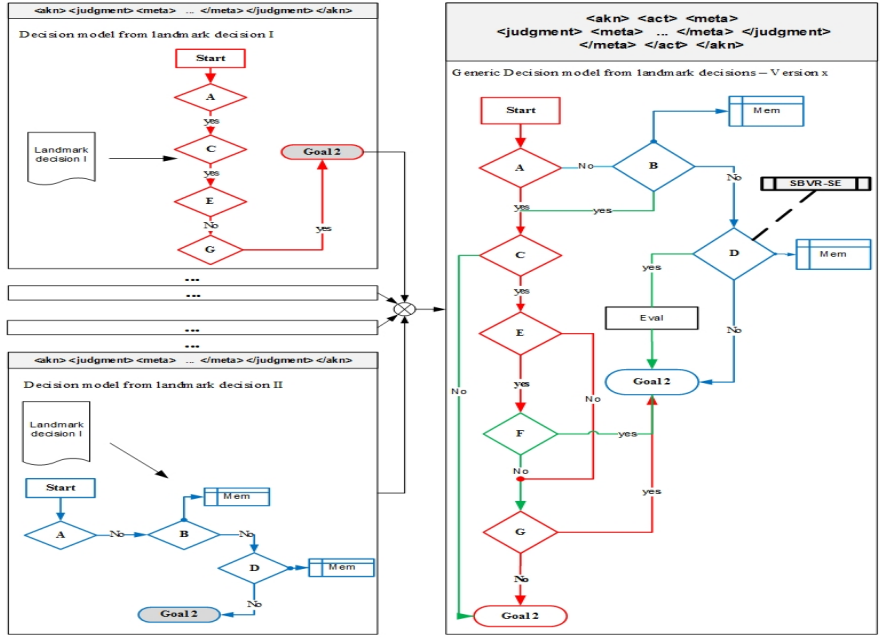

Fig. 2: Decision model for a legal section under consideration (adapted from [21]).

Further, we use the easy to understand decision model as basis for writing the legal norms and their elementary concerns in terms of constitutive vocabulary definitions and prescriptive behavioral legal rules in SBVR's Structured English, thus forming the semi-formal legal norm representation. One such semi-formal representation of a legal (procedural) norm from the decision point 'D' (from Fig 2) is as shown below:

- It is obligatory that the office_action *includes* **Paragraph_7_33_01**, if claim *is_rejected_under* essential_subject_matter_requirement.

The semi-formal procedural rules are built on legal facts, and **legal facts** are built on **legal concepts** [5] which are expressed by **legal terms**. Annotation of legal concept with the meta-data description (as shown in Fig 3) enables identification and representation of associated context information. A more in-depth discussions pertaining to SBVR and its use in legal domain has been presented by us in [22][23].

Further, for formal rule representation, we use KR4IPLaw - a patent norm representation format, which seamlessly integrates into the existing rule (including LegalRule) representation standards like RuleML [8], ReactionRuleML [24] and LegalRuleML [10]. Fig 4 depicts the general structure of KR4IPLaw.

---
[5] Represented using *green* color

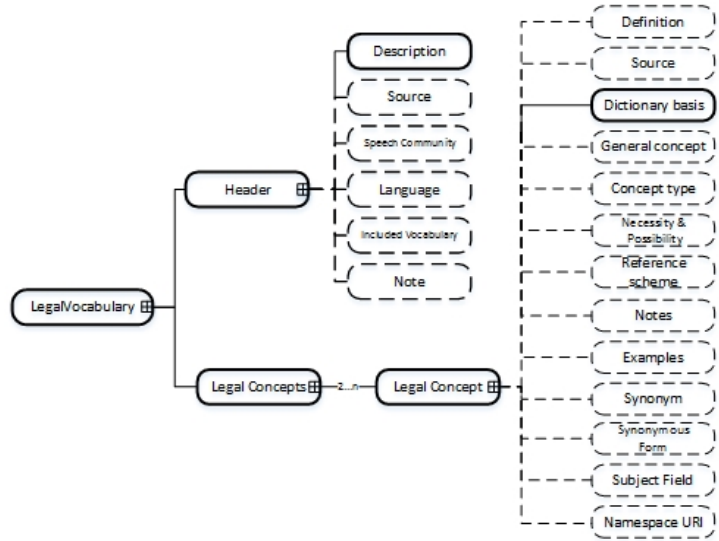

Fig. 3: Context Information pertaining to a Legal Concept.

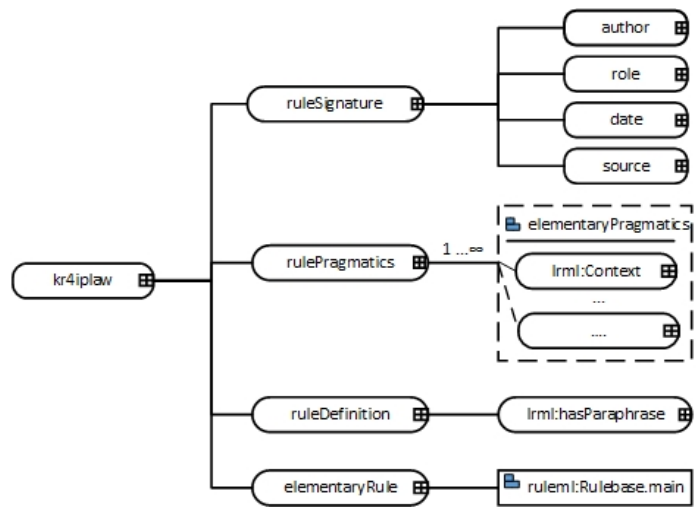

Fig. 4: KR4IPLaw: Rule Representation Format.

The module `<rulePragmatics>` holds all the pragmatic information which includes meta-information such as; `<Sources>` to identify the collection of legal resources relevant to the Legal document, `<References>` to provide an isomorphic relationship between the formal rule and the legally binding statements to the rule, `<Authority>`, `<Jurisdictions>`, `<Associations>`, `<TimeInstants>` etc.. While the general structure provided here depicts the intended semantics to the formal representation formats, the authors would like to refer to a series of publications for a detailed discussion on it [12][25][26].

Further, for reasoning, the formal rule is transformed into a Platform Specific Model (PSM) rule representation format, Prova [13] -a semantic web rule language and a high expressive distributed rule engine- where the pragmatic information from the PIM layer, associated to a norm are handled with by *guards*.

## 6 Conclusion and Future Directions

The paper initially discussed in detail the aspect of *Pragmatics* from different perspectives such as, computer science, Legal theory and speech-act theory. We then provided an cognitive approach for the interpretation of pragmatics by expanding the maxims. Finally, we discussed how such pragmatic information are captured (and transformed) at different layers of legal knowledge representations for its use in (semi-/)-automated legal norm reasoning.

## 7 Acknowledgement

This work has been partially supported by the "InnoProfile-Transfer Corporate Smart Content" project funded by the German Federal Ministry of Education and Research (BMBF) and the BMBF Innovation Initiative for the New German Länder - Entrepreneurial Regions.